\documentclass[10pt,twocolumn,letterpaper]{article}

\usepackage{cvpr}
\usepackage{times}
\usepackage{epsfig}
\usepackage{graphicx}
\usepackage{amsmath}
\usepackage{amssymb}
\usepackage[american]{babel}

\usepackage[breaklinks=true,bookmarks=false]{hyperref}

\cvprfinalcopy 

\def\cvprPaperID{****} 

\begin{document}

\title{ Revisiting the Effectiveness of Off-the-shelf Temporal Modeling \\ Approaches for Large-scale Video Classification}

\author{ Yunlong Bian, Chuang Gan$^{*}$ , Xiao Liu, Fu Li, Xiang Long \\
Yandong Li, Heng Qi, Jie Zhou, Shilei Wen, Yuanqing Lin\\
Baidu IDL \& Tsinghua University\\
}

\maketitle

\renewcommand{\thefootnote}{\fnsymbol{footnote}}
\footnotetext[1]{Corresponding author.}
\begin{abstract}
  This paper describes our solution for the video recognition task of ActivityNet Kinetics challenge that ranked the 1st place. Most of existing state-of-the-art video recognition approaches are in favor of an end-to-end pipeline. One exception is the framework of DevNet~\cite{devnet}. The merit of DevNet is that they first use the video data to learn a network (\ie fine-tuning or training from scratch). Instead of directly using the end-to-end classification scores (\eg softmax scores), they extract the features from the learned network and then fed them into the off-the-shelf machine learning models to conduct video classification. However, the effectiveness of this line work has long-term been ignored and underestimated. In this submission, we extensively use this strategy. Particularly, we investigate four temporal modeling approaches using the learned features: Multi-group Shifting Attention Network, Temporal Xception Network, Multi-stream sequence Model and Fast-Forward Sequence Model. Experiment results on the challenging Kinetics dataset demonstrate that our proposed temporal modeling approaches can significantly improve existing approaches in the large-scale video recognition tasks.  Most remarkably, our best single Multi-group Shifting Attention Network can achieve 77.7\% in term of top-1 accuracy and 93.2\% in term of top-5 accuracy on the validation set.
\end{abstract}

\section{Introduction}
Video understanding is among one of the most fundamental research problems in computer vision and machine learning. The ubiquitous video acquisition devices (e.g., smart phones, surveillance cameras, etc.) have created videos far surpassing what we can watch. It has therefore been a pressing need to develop automatic video understanding and analysis algorithms for various applications.

To recognize actions and events in videos, recent approaches based on deep convolutional neural networks (CNNs)~\cite{Sports1M,Twostream,devnet,C3D,gan2016you} and/or recurrent networks~\cite{LSTM,ULSTM,cho2014properties} have achieved state-of-the-art results. However, due to the lack of public available datasets, existing video recognition approaches are restricted to understand small-scale data, while large-scale video understanding remains an under-addressed problem.  To remedy this issue, Google DeepMind releases a new large-scale video dataset, named as Kinetics dataset~\cite{kay2017kinetics}, which contains 300K video clips of 400 human action class.

To address this challenge, our solution follows the strategy of DevNet framework~\cite{devnet}. Particularly, we first learn the basic RGB, Flow and Audio neutral network models using the videos. Then we extract the multi modality feature and fed them into different off-shelf temporal models. We also design four novel temporal modeling approaches, namely Multi-group Shifting Attention Network, Temporal Xception Network, Multi-stream sequence Model and Fast-Forward Sequence Model. Experiment results verity the effectiveness of the four models over the traditional temporal modeling approaches. We also find that these four temporal modeling approaches are complementary with each others and lead to the state-of-the-arts performances after ensemble.

The remaining sections are organized as follows. Section \ref{sec:2} presents the basic multi modal feature extraction. Section \ref{sec:3} describe our proposed off-shelf temporal modeling approaches. Section~\ref{sec:4} reports empirical results, followed by discussions and conclusions in Section \ref{sec:5}.

\begin{figure*}[t!]
    \centering
    \includegraphics[width=0.95\textwidth]{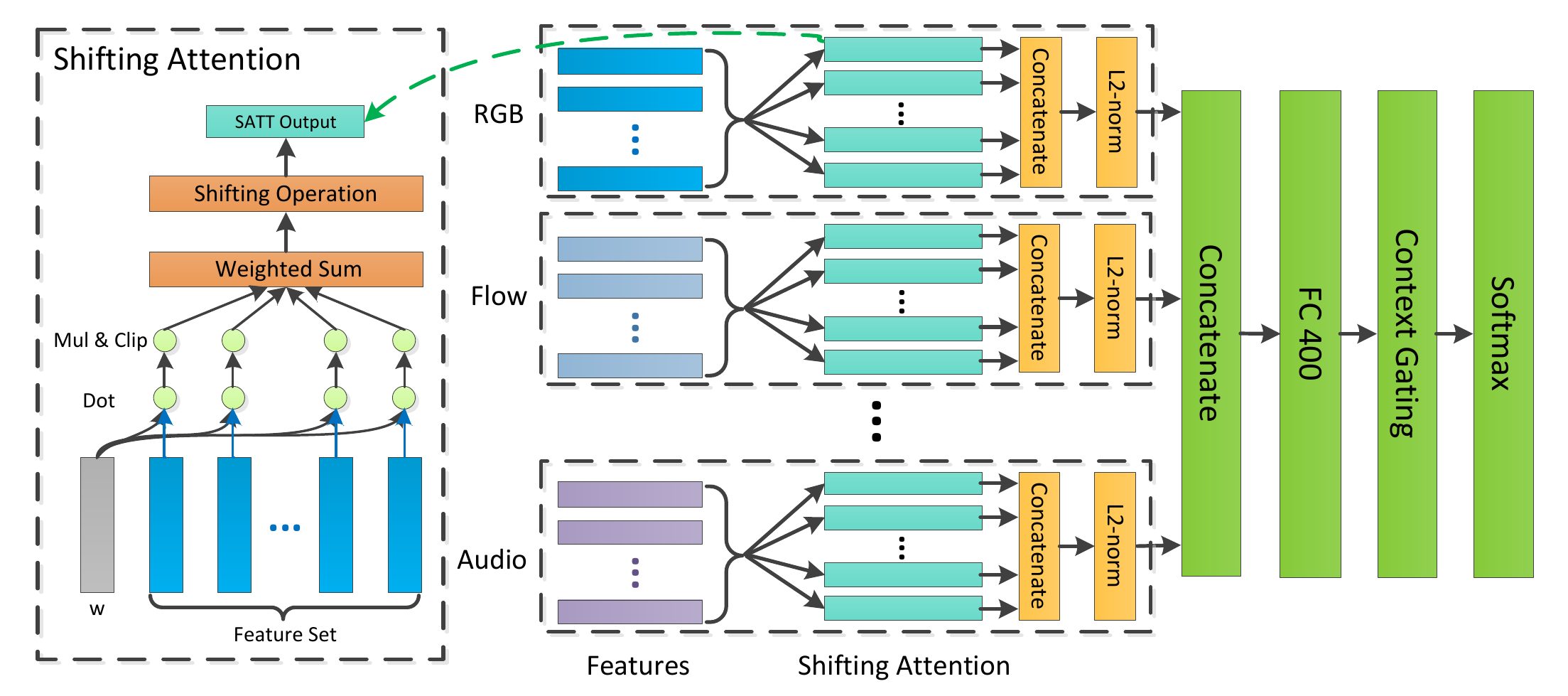}\\
    \caption{Multi-group Shifting Attention Network.}
    \label{fig:satt}
\end{figure*}

\section{Multimodal Feature Extraction}

\label{sec:2}

Videos are naturally multimodal because a video can be decomposed into visual and acoustic components, and the visual component can be further divided into spatial and temporal parts. We extracted multi modal features to best represent videos accordingly.
\subsection{Visual Feature}
As in \cite{Twostream}, we used RGB images for spatial feature extraction and stacked optical flow fields for temporal feature extraction. We tried different ConvNet architectures and found Inception-ResNet-v2 \cite{inception4} outperforms others in both spatial and temporal components. The RGB model is initialized with pre-trained model from ImageNet and fine-tuned in the Kinetics dataset, while the flow model is initialized from the fine-tuned RGB model. Inspired by \cite{WangX16}, the temporal segment network framework is used and three segments are sampled from each trimmed video for video-level training. During testing, we can densely extract features for each frames in the video.

\subsection{Acoustic Feature}
We use ConvNet-based audio classification system \cite{audiocnn} to extract acoustic feature. The audio is divided into 960ms frames, and the frames are processed with Fourier transformation, histogram integration and logarithm transformation. The resulting frame can be seen as a ${96\times64}$ image that form the input of a VGG16 \cite{Simonyan14c} image classification model. Similar with the visual feature, we trained the acoustic feature in the temporal segment network framework.

\section{Off-shelf Temporal Modeling Approaches}

\label{sec:3}
 In this section, we present a brief introduction of our proposed shifting attention network and temporal Xception network.
 More implementation details and analysis will be in a following technique report. We also refer \cite{li2017temporal} for the details of multi-stream sequence model and fast-forward sequence model.

\subsection{Shifting Attention Network}
Attention models have shown great potential in sequence modeling. For example, numerous pure attention architectures \cite{AttentionIsAllYouNeed, SelfAttentive} have been proposed and achieved promising results in natural language processing problems. In order to explore the capabilities of attention models in action recognition, a shifting attention network architecture is proposed, which is efficient, elegant and solely based on attention.

\subsubsection{Shifting Attention}
An attention function can be considered as mapping a set of input features to a single output, where the input and output are both matrices that concatenate feature vectors. The output of the shifting attention $\textrm{SATT}(X)$ is calculated through a shifting operation based on a weighted sum of the features:

\begin{equation}
	\textrm{SATT}(X)  = \frac{\lambda X \cdot a + b}{\left \lVert \lambda X \cdot a + b \right \rVert_2},
\end{equation}
where $\lambda$ is a weight vector calculated as

\begin{equation}
\lambda = \textrm{softmax} (\alpha \cdot w X^T),
\end{equation}
$w$ is learnable vector, $a$ and $b$ are learnable scalars, and $\alpha$ is a hyper-parameter to control the sharpness of the distribution. The shifting operation actually shifts the weighted sum and at the same time ensures scale-invariance. The shift operation efficiently enables different attention components to flexibly diverge from each other and have different distributions. This lays the foundation for Multi-SATT, which we describe next.

\begin{figure*}[t!]
    \centering
    \includegraphics[width=0.95\linewidth]{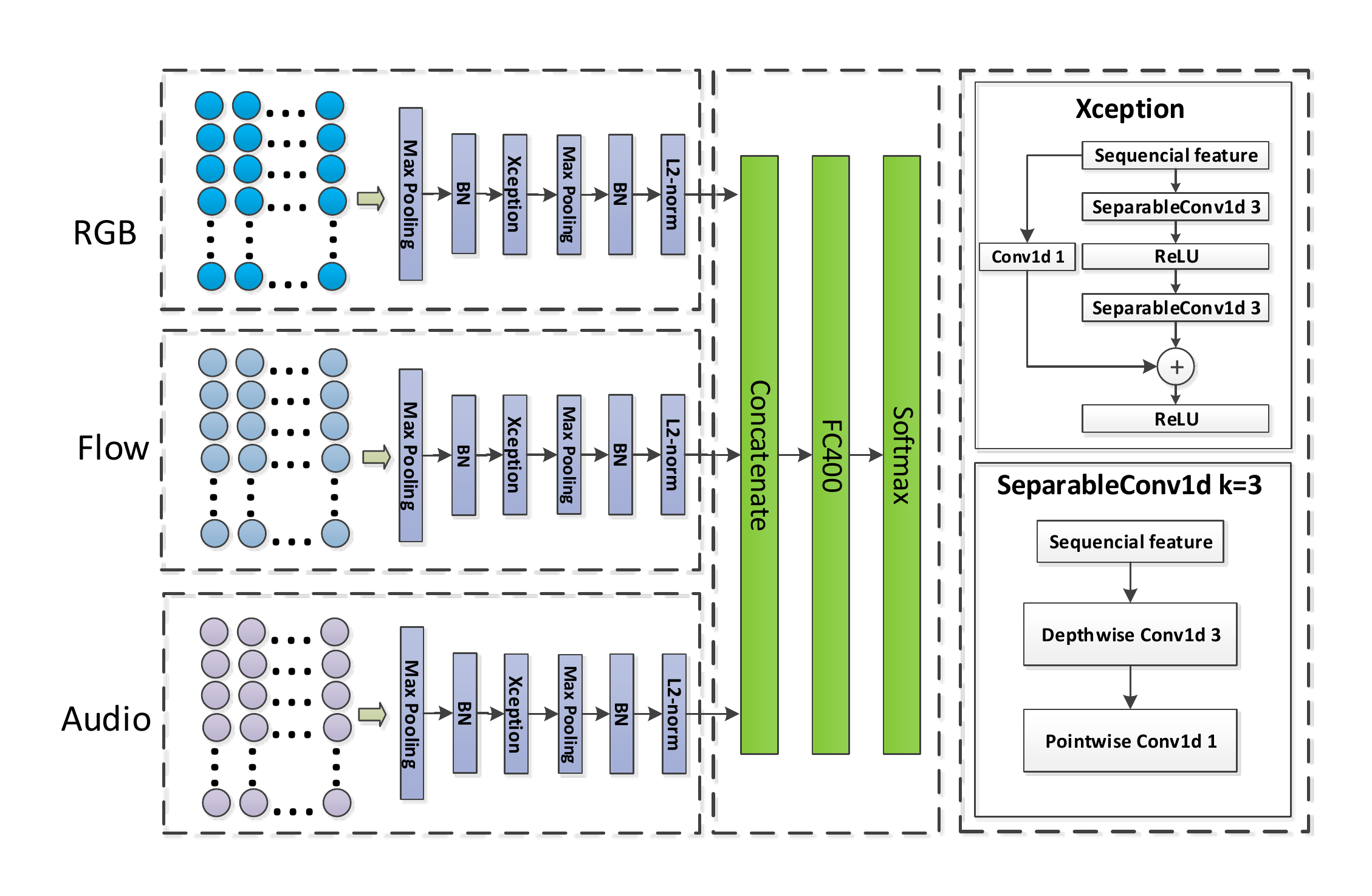}\\
    \caption{Temporal Xception Network.}
    \label{fig:xception}
\end{figure*}

\subsubsection{Multi-Group Shifting Attention Network}
In order to collect multi modal information from videos, we extract a variety of different features, such as appearance (RGB), motion (flow) and audio signals. Although the attention model focuses on some specific features and effectively filters out irrelevant noise, it is unrealistic to merge all multi modal feature sets within one attention model, because features of different modality have different values, dimensions and scales.
Instead, we propose Multi-Group Shifting Attention Networks for training multiple groups of attentions simultaneously.
The architecture of the proposed Multi-SATT is illustrated in Figure \ref{fig:satt}.

First, we extract multiple feature sets from the video. For each feature set $X_i$, we apply $N_i$ different shifting attentions, which we call one attention group, and then we concatenate the outputs. Next, the outputs of different attention groups are normalized separately and concatenated to form a global representation vector for the video. Finally, the representation vector is used for classification through a fully-connected layer.

\subsection{Temporal Xception Network}
Depthwise separable convolution architecture~\cite{chollet2016xception,xie2016aggregated} has shown its power in image classification by reducing the number of parameters and increasing classification accuracy simultaneously. Recently, convolutional sequence-to-sequence networks have been successfully applied to machine translation tasks~\cite{gehring2017convolutional,kaiser2017depthwise}. In this competition, we adopt the temporal Xception network for action recognition, which apply the depthwise separable convolution families to the temporal dimension and achieves promising performance.
The proposed temporal Xception network architecture is shown in Figure \ref{fig:xception}. Zero-valued multi modal features were padded to make fixed length data for each stream. We applied adaptive temporal max pooling to obtain $n$ segments for each video. We then feed the video segment features into a Temporal Convolutional block, which is consist of a stack of two separable convolutional layers followed by batch norm and activation with a shortcut connection. Finally, the outputs of three stream features are concatenated and fed into the fully-connected layer for classification.

\begin{table*}[t]
  \centering
  \begin{tabular}{c|c|c|c}
  \hline
  \hline
  Model & Modality & Top-1 Accuracy (\%) & Top-5 Accuracy (\%)\\
  \hline
   Inception-ResNet-v2&RGB & 73.0 & 90.9 \\
  \hline
    Inception-ResNet-v2& Flow & 54.5 & 75.9 \\
  \hline
    VGG16& Audio & 21.6 & 39.4 \\
     \hline
    Late fusion & RGB + Flow + Audio & 74.9 & 91.6 \\
  \hline
  \hline
  Multi-stream Sequence Model &RGB + Flow + Audio & 77.0 & 93.2 \\
  \hline
  Fast-forward LSTM &RGB + Flow + Audio & 77.1 & 93.2 \\
  \hline
  Temporal Xception Network &RGB + Flow + Audio& 77.2 & \textbf{93.4} \\
  \hline
  Shifting Attention Network & RGB + Flow + Audio& \textbf{77.7} & 93.2 \\
  \hline
    \hline
  Ensemble & RGB + Flow + Audio & \textbf{81.5} & \textbf{95.6} \\
  \hline
  \hline
  \end{tabular}
  \caption{Kinetics validation results.}
  \label{table:result}
\end{table*}

\section{Experiment Results}
\label{sec:4}
We conduct experiment on the challenging Kinetics dataset
The dataset contains 246,535 training videos, 19,907 validation videos and 38,685 testing videos. Each video is in one of 400 categories.

Table \ref{table:result} summarizes our results on the Kinetics validation dataset. From Table~\ref{table:result}, we have three key observations. (1) Temporal modeling approaches with multi modal features are a more effective approach than naive combining the classification scores of different modality networks for the video classification. (2) The proposed Shifting Attention Network and Temporal Xception Network can achieve comparable or even better results than the traditional sequence models (e.g. LSTM), which indicates they might serve as alternative temporal modeling approaches in future. (3) Different temporal modeling approaches are complementary to each other.

\section{Conclusions}
\label{sec:5}
In this work, we have proposed four temporal modeling approaches to address the challenging large-scale video recognition task.
Experiment results verify that our approaches achieve significantly better results than the traditional temporal pooling approaches.
The ensemble of our individual models has been shown to improve the performance further, enabling our method to rank first worldwide in the challenge competition. All the code and models will be released soon.

\end{document}